\documentclass[letterpaper, 10 pt, conference]{ieeeconf}

\IEEEoverridecommandlockouts                             

\usepackage{graphicx}
\usepackage{multicol}
\usepackage{booktabs}
\usepackage{amsmath,amssymb}
\usepackage[utf8]{inputenc}
\usepackage[english]{babel}
\usepackage{multirow}
\usepackage[font=small]{caption}
\usepackage{float}
\usepackage[hidelinks]{hyperref}

\DeclareMathOperator*{\argmin}{argmin}
\usepackage{lettrine}
\usepackage[]{algorithm2e}
\usepackage{algpseudocode}
\usepackage{cite}
\usepackage{filecontents}
\usepackage{lipsum}
\usepackage{color}
\usepackage{esdiff}
\usepackage{epstopdf}
\usepackage[normalem]{ulem}
\usepackage{soul}

\usepackage{subcaption}
\usepackage{xcolor}
\usepackage{colortbl}
\usepackage{balance}
\newtheorem{defn}{Definition}[section]

\definecolor{pink}{rgb}{1, 0, 1}
\definecolor{orange}{rgb}{1, 0.7529, 0}
\definecolor{darkgreen}{rgb}{0, 0.8, 0}
\begin{document}

\title{CAP: A Connectivity-Aware Hierarchical Coverage Path Planning Algorithm for Unknown Environments using Coverage Guidance Graph}

\author{Zongyuan Shen, Burhanuddin Shirose, Prasanna Sriganesh and Matthew Travers% <-this % stops a space
\thanks {All authors are from the Robotics Institute, Carnegie Mellon University, USA. {\{zongyuas, bshirose, pkettava, mtravers\}@andrew.cmu.edu}}
\thanks{Supplementary video - \scriptsize\texttt{\url{https://youtu.be/1pH-PkcRVZg}}}
}
        
\maketitle

\begin{abstract}
Efficient coverage of unknown environments requires robots to adapt their paths in real time based on on-board sensor data. In this paper, we introduce CAP, a connectivity-aware hierarchical coverage path planning algorithm for efficient coverage of unknown environments. During online operation, CAP incrementally constructs a coverage guidance graph to capture essential information about the environment. Based on the updated graph, the hierarchical planner determines an efficient path to maximize global coverage efficiency and minimize local coverage time. The performance of CAP is evaluated and compared with five baseline algorithms through high-fidelity simulations as well as robot experiments. Our results show that CAP yields significant improvements in coverage time, path length, and path overlap ratio.
\end{abstract}

\section{Introduction}
\begin{figure}[t]
        \centering
    \subfloat[Greedy method.]{
        \includegraphics[width=0.48\columnwidth]{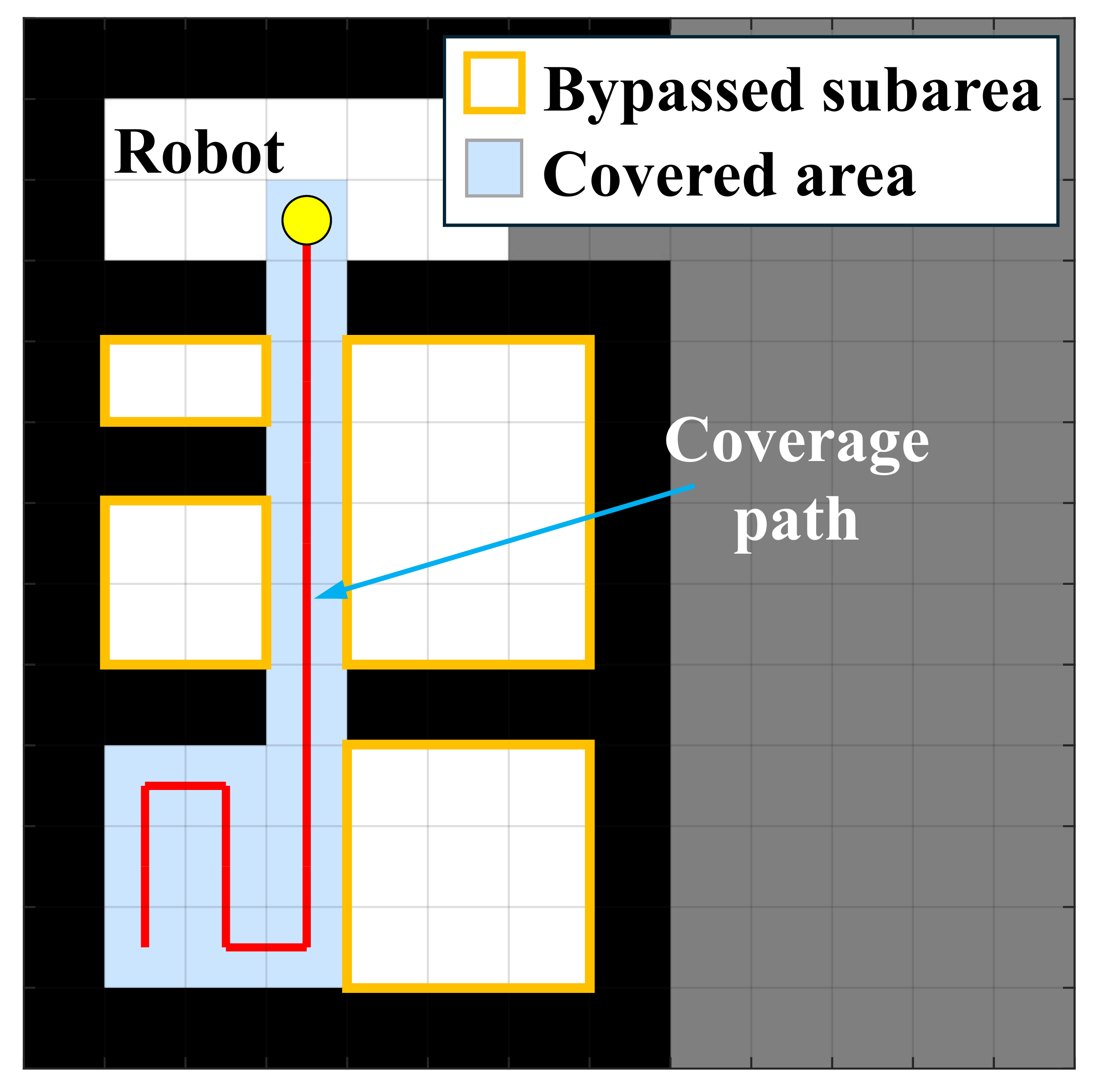}\label{fig:greedyStrategy_part1}}\quad \hspace{-12pt}
        \centering
    \subfloat[Our proposed method (CAP).]{
        \includegraphics[width=0.48\columnwidth]{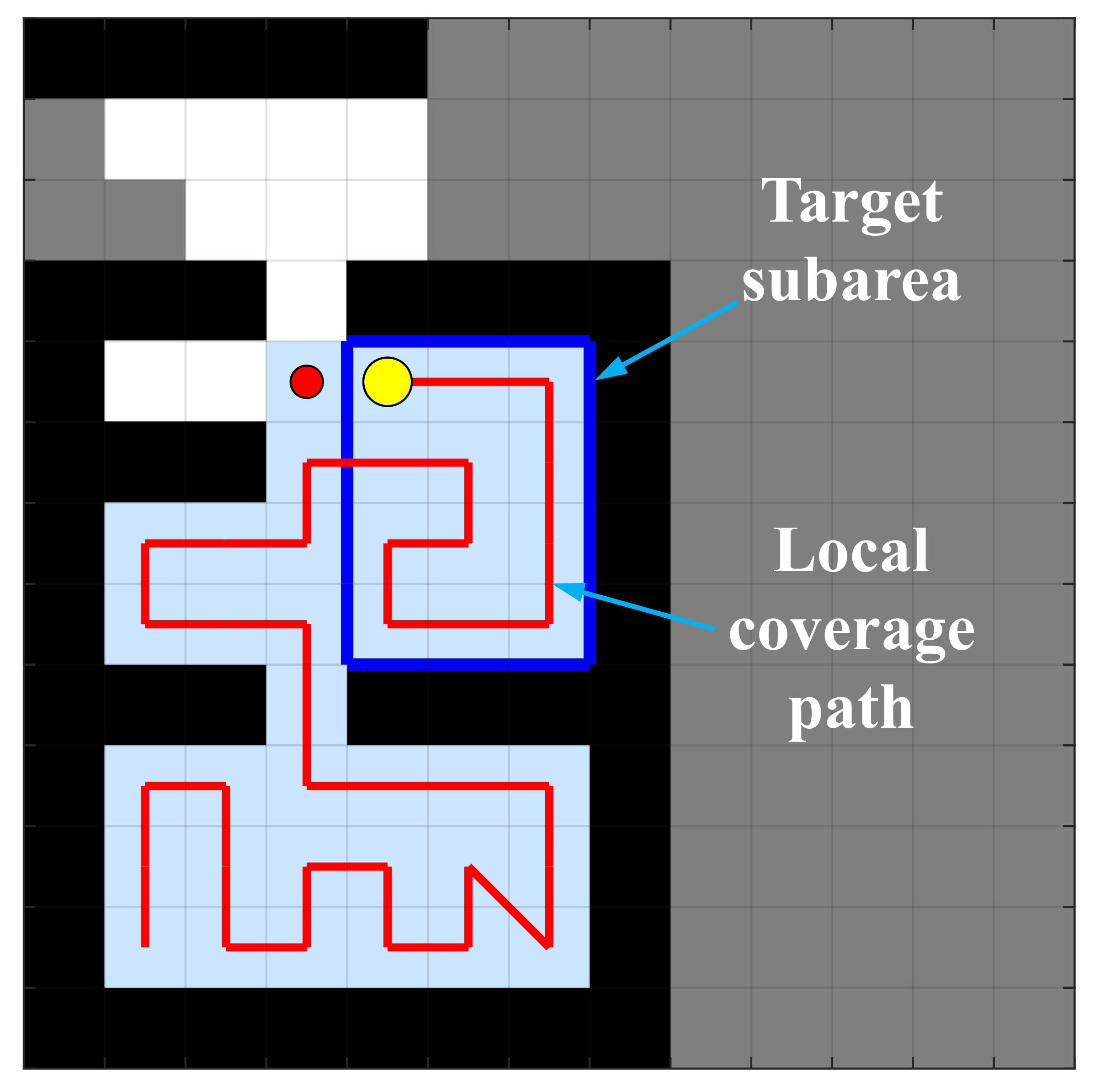}\label{fig:ours}}\\
    \caption{Example of coverage paths generated by different methods.}\label{fig:greedyStrategy} 
     \vspace{-1.5em}
 \end{figure}

Optimized coverage path planning (CPP) enables robots to achieve complete coverage of all points in a search area efficiently. Existing CPP methods generate well-defined coverage patterns (e.g., the back-and-forth pattern shown in Fig.~\ref{fig:greedyStrategy_part1}) and ensure complete coverage in finite time, but prioritize immediate cost minimization, not considering a global perspective of the entire search area. These myopic decisions result in inefficient global coverage paths. Enabling efficient coverage requires reasoning about the topology of known and unknown parts of the environment and adapting the robot's path as new information becomes available. To this end, we present a hierarchical coverage path planning algorithm that utilizes essential global information about the environment to efficiently guide the CPP process online.

To enable efficient coverage of unknown environments, we introduce \textbf{CAP}, a \textbf{C}onnectivity-\textbf{A}ware Hierarchical Coverage Path \textbf{P}lanning algorithm. CAP maintains and updates an environmental map online that makes it possible to dynamically identify disconnected subareas of an environment. The subareas are split into \textit{exploring} (free-space adjacent to unknown regions) and \textit{explored} (free-space fully mapped by the sensor, not adjacent to unknown regions) areas. CAP then incrementally constructs a coverage guidance graph by adding these subareas as new nodes with edges representing local connectivity between subareas, capturing the global topological structure of the environment. The graph is then used to compute an optimal subarea traversal tour by solving a traveling salesman problem (TSP). The TSP-based tour ensures that the robot follows a globally efficient sequence of subareas to explore.  We find that this globally aware planner addresses issues of bypassing isolated regions and inefficient backtracking (Fig. \ref{fig:greedyStrategy}). The coverage planner dynamically switches between a greedy strategy to initially cover the \textit{exploring} subareas and a TSP-based optimal coverage plan within the \textit{explored} subareas to further minimize local coverage time. CAP is illustrated in Fig.~\ref{fig:example}. 

The specific contributions of this work are: 
\begin{itemize}
\item Development of CAP for efficient coverage of unknown environments with two key components:
\begin{itemize}
    \item Coverage guidance graph to  capture essential environmental information for efficient planning.
    \item Hierarchical coverage path planning method to improve the global coverage efficiency and minimize local coverage time.
\end{itemize}
\item Comprehensive comparison to baseline algorithms in simulation and real-world experiments.
\end{itemize}

\section{Related Work}\label{sec:review}
Coverage path planning has been utilized in a variety of real-world robotic applications, such as seabed mapping, structural inspection, spray painting surfaces, floor cleaning, and aerial surveying~\cite{ shen2022ct,song2020online, clark2024online}. CPP methods are broadly classified as offline or online~\cite{galceran2013survey}. Offline methods \cite{cao2020hierarchical} rely on prior information of the environments to design coverage paths, which can be highly effective in ideal conditions. However, their performance degrades if the prior information is incomplete and/or incorrect. Therefore, it is essential to develop online CPP methods, which adapt the coverage path in real-time based on the collected environmental information.
 
Acar and Choset~\cite{acar2002sensor} developed a method which detects the critical points to decompose the area into multiple cells and incrementally constructs a Reeb graph that has edges as the cells and nodes as the critical points. Each cell is covered by certain coverage pattern. Gabriely and Rimon~\cite{gabriely2003competitive} developed the full spiral spanning tree covering algorithm to generate a spiral coverage path by following a spanning tree. Gonzalez et al.~\cite{gonzalez2005bsa} presented the backtracking spiral algorithm (BSA), which generates a
spiral path by following the uncovered area boundary and switches to a backtracking mechanism for escaping from the dead-end situation. 

Luo and Yang~\cite{luo2008bioinspired} presented the biologically-inspired neural network (BINN) algorithm, which places a neuron at each grid cell. An activity landscape is updated such that the obstacle cells have negative potentials for collision avoidance and uncovered cells have positive potentials for coverage. Then, the target cell is selected in the local vicinity of robot using this activity landscape. Later in~\cite{huo2024}, this algorithm was improved to mitigate the dead-end situation. Viet et al.~\cite{viet2013ba} presented the boustrophedon motions and A$^*$ (BA$^*$) algorithm, which performs back-and-forth motion until no uncovered cell is found in the local vicinity. Then, the robot moves to the backtracking cell to resume the coverage.

Song and Gupta~\cite{song2018} developed the $\epsilon^*$ algorithm, which utilizes an exploratory Turing Machine as a supervisor to guide the robot online with tasking commands for coverage via back-and-forth motion. The algorithm uses multi-resolution adaptive potential surfaces to escape from a dead-end situation. Hassan and Liu~\cite{hassan2019ppcpp} proposed the predator–prey CPP (PPCPP) algorithm, which determines target cell in the local vicinity of robot according to the immediate reward consisting of three components: 1) predation avoidance reward that maximizes the distance to the predator (predator is a user-defined stationary point), 2) smoothness reward that minimizes the turn angle, and 3) boundary reward for covering boundary of the uncovered area. 

Some CPP methods have also been developed for 3D structure coverage \cite{shen2022ct,song2020online}, multi-robot systems~\cite{zhang2024herd,clark2024online}, curvature-constrained robots~\cite{shen2019online,maini2022online}, tethered robots~\cite{peng2025,shnaps2014}, and energy-constrained robots~\cite{shen2020,dogru2022eco}.

\section{CAP Algorithm}\label{sec:algorithm}
 
\begin{figure*}[t]
        \centering
    \subfloat[Target cell is picked by the  greedy strategy in the local vicinity.]{
        \includegraphics[width=0.48\columnwidth]{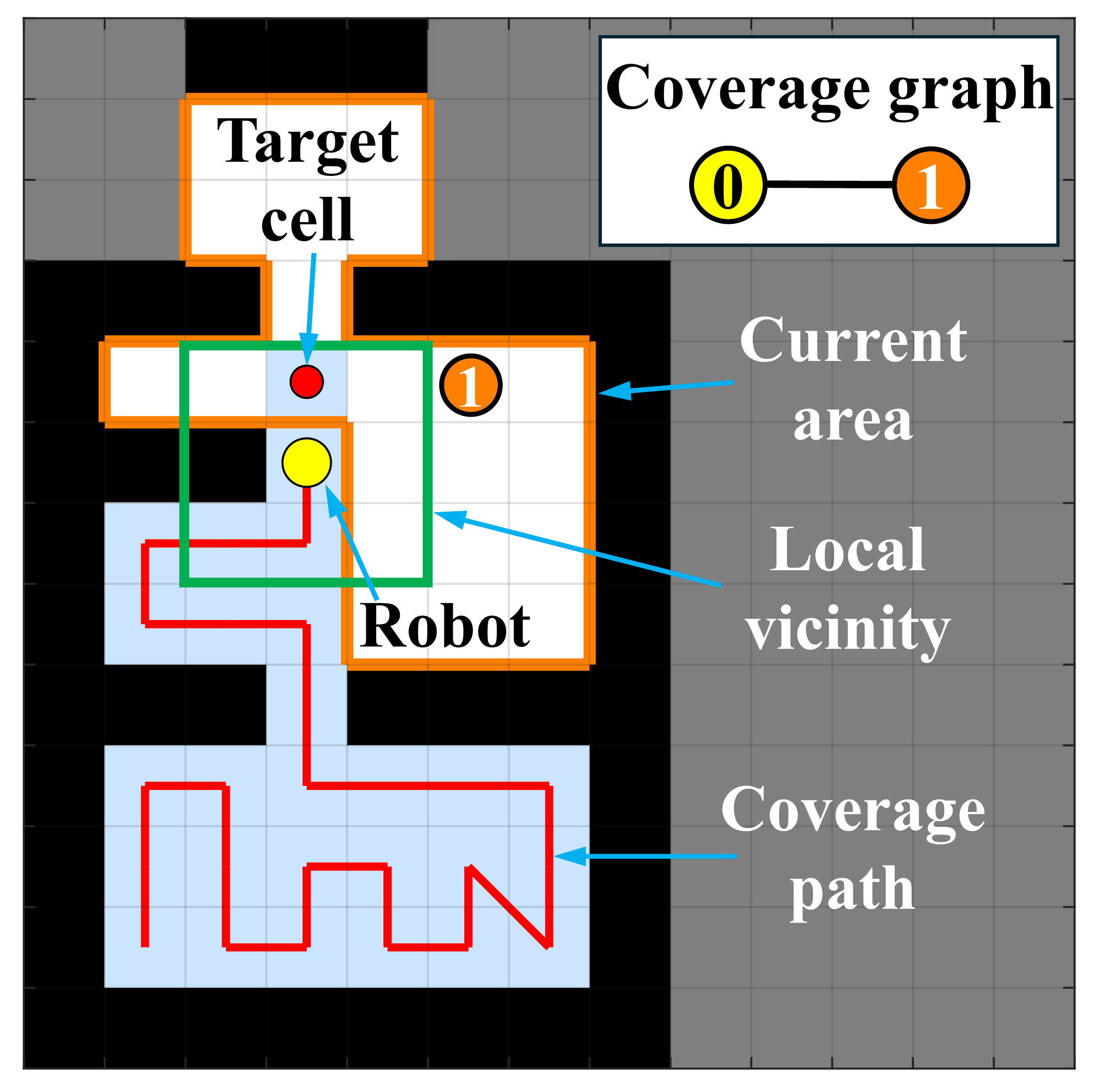}\label{fig:example_part1}}\quad \hspace{-10pt}
        \centering
    \subfloat[Three disconnected subareas are identified and categorized.]{
        \includegraphics[width=0.48\columnwidth]{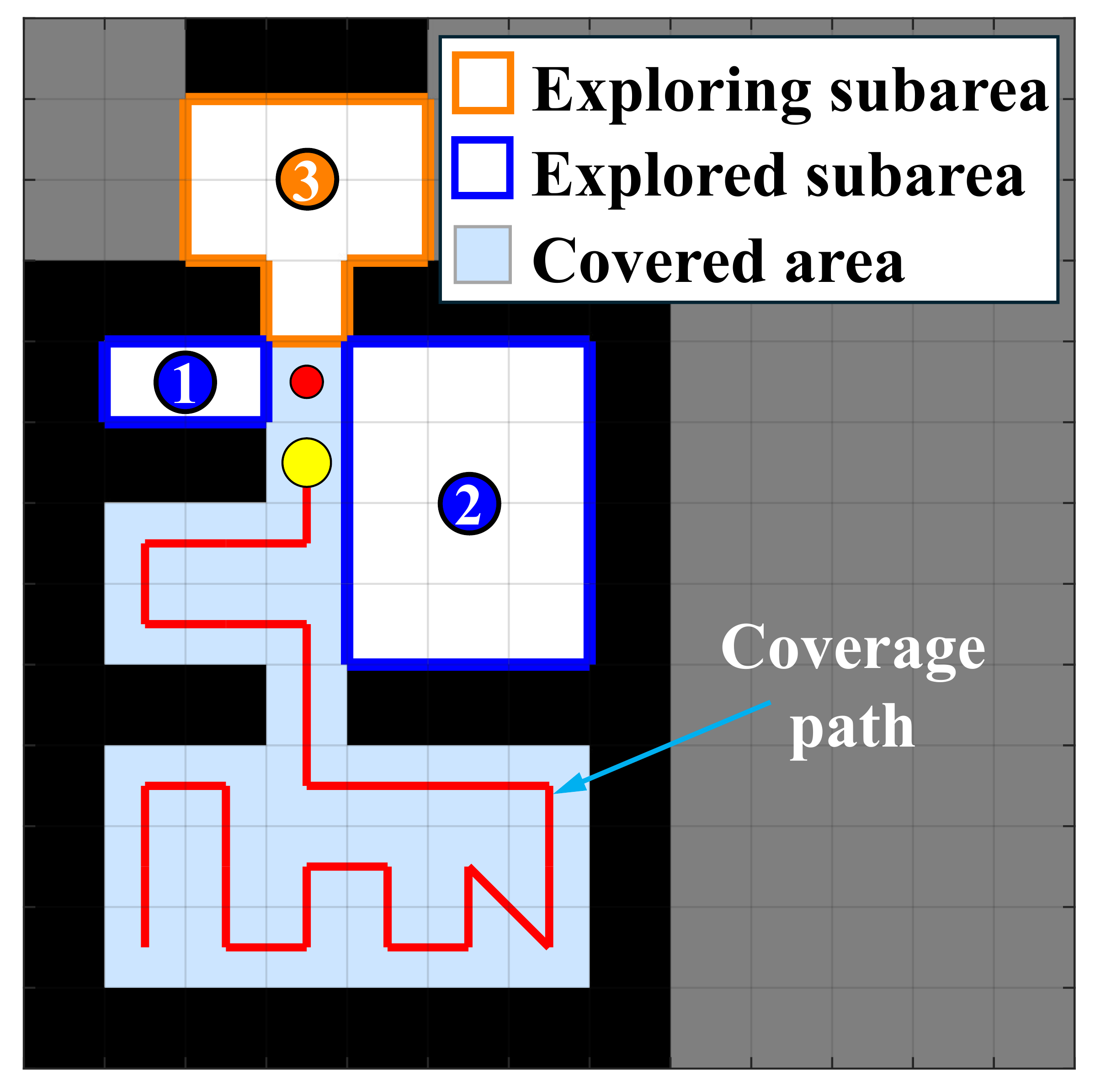}\label{fig:example_part2}}\quad \hspace{-10pt}
        \centering
    \subfloat[Coverage guidance graph is updated and global tour is computed.]{
        \includegraphics[width=0.48\columnwidth]{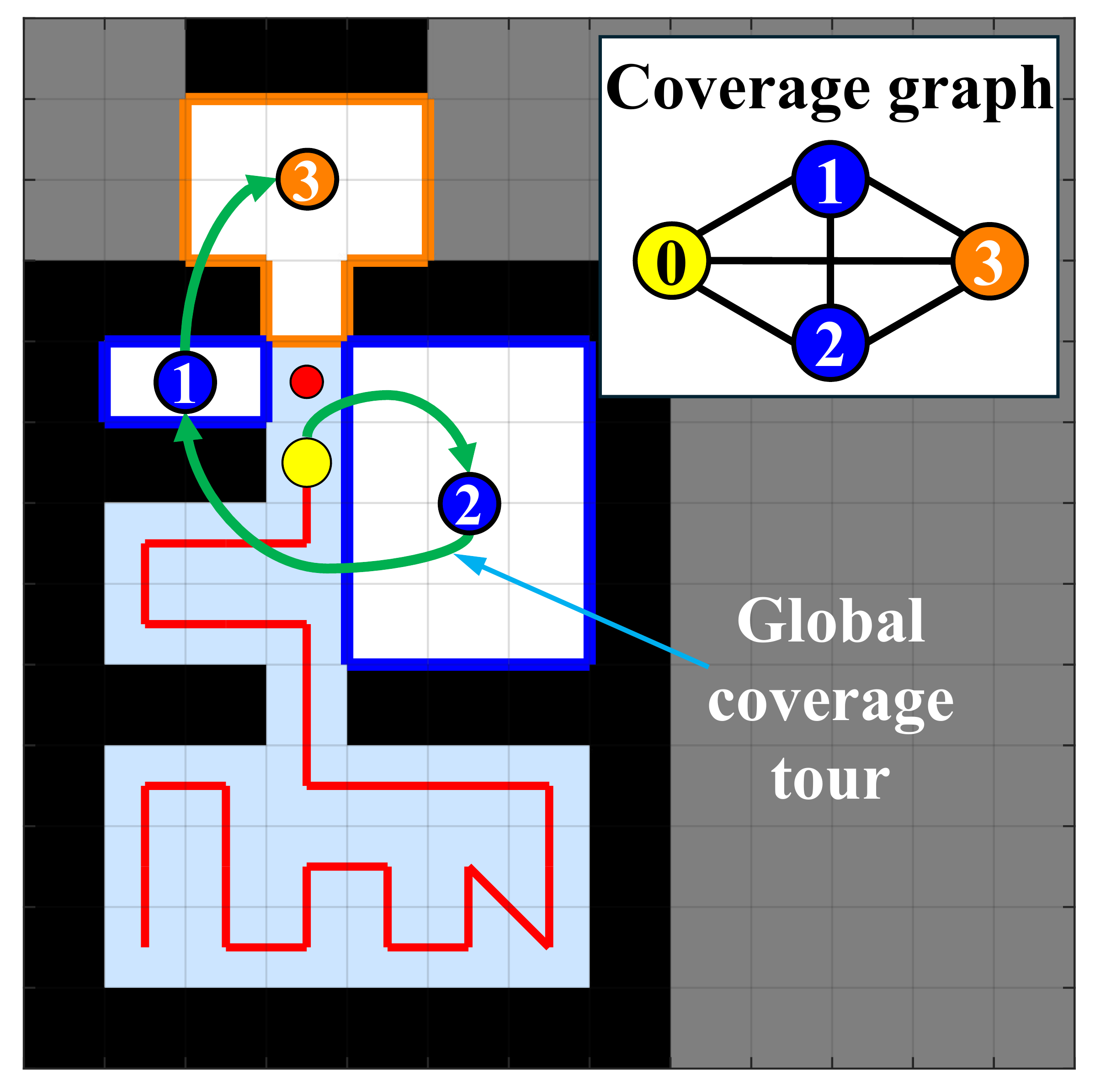}\label{fig:example_part3}}\quad \hspace{-10pt}
        \centering
    \subfloat[Local coverage path is generated to completely cover target subarea.]{
        \includegraphics[width=0.48\columnwidth]{example_part4.pdf}\label{fig:example_part4}}\\
    \caption{Illustration of the CAP algorithm: a)-b) identification of disconnected subareas, c) incremental construction of coverage guidance graph, and c)-d) computation of hierarchical coverage path.}\label{fig:example} 
    \vspace{-1.5em}
 \end{figure*}

Let $\mathcal{A}\subset\mathbb{R}^2$ be the unknown area populated by obstacles. First, a tiling is constructed on $\mathcal{A}$ as defined below.

\begin{defn}[Tiling]\label{define:Tiling}
A set $\mathcal{T} = \{\tau_{j} \subset \mathbb{R}^2:j= 1,\ldots,|\mathcal{T}|\}$ is a tiling of $\mathcal{A}$ if its elements 1) are mutually exclusive, i.e., $\tau_{j} \cap \tau_{j'} =\emptyset, \forall {j},{j'} \in \{1,\ldots,|\mathcal{T}|\}, {j} \neq {j'}$, and 2) cover $\mathcal{A}$, i.e., $\mathcal{A} \subseteq \bigcup_{j=1}^{|\mathcal{T}|}\tau_{j}$.
\end{defn}

It is recommended that the cell is big enough to contain the robot and small enough for the tasking device/sensor (e.g., cleaning brush) to completely cover it when the robot visits it. During online operation, the robot updates environmental information on $\mathcal{T}$ based on onboard sensor data and picks an uncovered free cell to cover. Thus, $\mathcal{T}$ is partitioned into three sets: 1) obstacle ($\mathcal{T}_o$), 2) free ($\mathcal{T}_f$), and 3) unknown ($\mathcal{T}_u$). The free cells are further separated into two subsets: covered ($\mathcal{T}_f^c$) and uncovered ($\mathcal{T}_f^u$). 

Let $t_c\in \mathbb{R}^+$ be the time instant at which the environmental map is updated and the objective of CAP is to find an updated path $\sigma:[t_c,T] \rightarrow \mathcal{T}_{f}$ to provide complete coverage of $\mathcal{A}$ with minimized coverage time, where $T\in \mathbb{R}^+$ is the total coverage time.

\begin{defn}[Complete Coverage] Let $\tau(t) \in \mathcal{T}_f$ be the cell visited by the robot at time $t\in[t_c,T]$. Then, the coverage of $\mathcal{A}$ is said to be complete if $\exists T\in \mathbb{R}^+$, s.t. the path $\sigma\triangleq[\tau(t)]_{t=t_c}^{T}$ covers $\mathcal{T}_f^u$, i.e., 

\begin{equation}
\mathcal{T}_f^u \subseteq \bigcup_{t\in[t_c,T]} \tau(t)
\end{equation}
\end{defn}

Therefore, we presents an algorithm consisting of three steps: 1) subarea identification, 2) coverage guidance graph construction, and 3) hierarchical coverage path planning.

\subsection{Identification of Disconnected Subareas}
\label{detectSubarea}

\begin{defn}[Exploring Area]\label{define:ExploringArea} An uncovered free area is said to be \textit{exploring} if it is adjacent to unknown area.
\end{defn}

As seen in Fig.~\ref{fig:example_part1}, the current area is an \textit{exploring} area. Thus, the robot follows the greedy strategy by selecting an uncovered cell as target cell $\tau_{target}$ in the local vicinity. This strategy could break the current area into multiple disconnected subareas that are surrounded by obstacles and covered area, as seen in Fig.~\ref{fig:example_part2}. 

\begin{defn}[Disconnected Subareas]\label{define:Disconnectedness} On tiling $\mathcal{T}$, two subareas are said to be disconnected if there is no path connecting them by only traversing uncovered area $\mathcal{T}_f^u$.
\end{defn}

The process to find disconnected subareas is described as follows. First, an uncovered cell which does not belong to any identified subarea is picked in the local vicinity of robot and $\tau_{target}$, and assigned an index $\ell \in \mathbb{N}^+$. Then, starting at this cell, the algorithm recursively searches and labels adjacent uncovered cells with the same index $\ell$ in four directions (i.e., up, down, left, right) until no more adjacent uncovered cells can be found. These labeled cells are grouped together to form a subarea $\mathcal{A}_{\ell}$. Its center position $p_\ell$ is computed as the average position of all its cells and then relocated into the nearest cell within $\mathcal{A}_{\ell}$ if it falls inside obstacle or unknown areas. After that, $\mathcal{A}_{\ell}$ is marked as \textit{exploring} if it is adjacent to unknown area. Otherwise, it is marked as \textit{explored}. The above process repeats to find all subareas $\{\mathcal{A}_\ell\}_{\ell=1}^{L}$. Fig. \ref{fig:example_part2} shows an example where three disconnected subareas are identified. These subareas are then added as new nodes to the coverage guidance graph.
\subsection{Incremental Construction of Coverage Guidance Graph} \label{updateGraph}

A coverage guidance graph is incrementally built based on the detected subarea information, and used to track the search progress and compute the hierarchical coverage path.

\begin{defn}[Coverage Guidance Graph]\label{define:coverage_graph}
A coverage guidance graph $\mathcal{G}=\left(\mathcal{N},\mathcal{E}\right)$ is defined as an undirected complete graph that consists of: 
\begin{itemize}
\item A node set $\mathcal{N} = \left\{n_\ell: \ell = 0,\ldots |\mathcal{N}|-1\right\}$, where $n_0$ represents robot's current position $p_{\mathcal{R}}$ and the remaining node $n_\ell$ corresponds to a subarea $\mathcal{A}_\ell$ with center at $p_\ell$.
\item An edge set $\mathcal{E}=\left\{e_\alpha: \alpha=0,\ldots\ |\mathcal{E}|-1\right\}$, where each edge $e_{\alpha}\equiv\left(n_i,n_j\right): n_i \neq n_j, \forall n_i,n_j \in \mathcal{N}$ represents the shortest collision-free path connecting two nodes computed by applying A$^*$ algorithm~\cite{hart1968formal} on tiling $\mathcal{T}$. 
\end{itemize}
\end{defn}

Fig.~\ref{fig:example} illustrates the process of incremental construction of the graph. It is first initialized with the node $n_0$ and $n_1$ which correspond to robot's current position $p_{\mathcal{R}}$ and the current area $\mathcal{A}_1$, respectively. As seen in Fig.~\ref{fig:example_part1}, the robot covers $\mathcal{A}_1$ using greedy strategy. The environmental map reveals that $\mathcal{A}_1$ is separated into three subareas, as shown in Fig.~\ref{fig:example_part2}. Then, the graph $\mathcal{G}$ is updated by pruning current node with its edge and adding detected subareas as new nodes with associated edges as shown in Fig.~\ref{fig:example_part3}. After that, a target node is selected from $\mathcal{G}$ for coverage. To track the coverage progress, each node of $\mathcal{G}$ is assigned a state using symbolic encoding $\Phi:\mathcal{N}\rightarrow \{C,\widetilde{O},\hat{O}\}$. While $C$ means that the node is already covered by the robot, $\widetilde{O}$ ($\hat{O}$) means that the node is open for coverage and its corresponding subarea is \textit{exploring} (\textit{explored}). This encoding generates the set partition $\mathcal{N}=\{\mathcal{N}_{C},\mathcal{N}_{\widetilde{O}},\mathcal{N}_{\hat{O}}\}$. Next, a global coverage tour is computed to cover all nodes in $\mathcal{N}_{\widetilde{O}}$ and $\mathcal{N}_{\hat{O}}$ as described below.

\subsection{Global Coverage Tour Planning}
\label{globalCover}

To find an optimal tour to cover all open nodes starting from robot's current position $p_{\mathcal{R}}$, a modified TSP-based strategy is proposed. First, a complete undirected graph $\bar{\mathcal{G}}=(\bar{\mathcal{N}},\bar{\mathcal{E}})$ is derived from $\mathcal{N}_{\widetilde{O}}$ and $\mathcal{N}_{\hat{O}}$ and consists of:

\begin{itemize}
\item A node set $\bar{\mathcal{N}} = \{\bar{n}_i: \bar{n}_0\equiv p_\mathcal{R} \ \textrm{and} \ \bar{n}_i\in \mathcal{N} \setminus \mathcal{N}_{C}, \ \forall i=1,\ldots|\bar{\mathcal{N}}|-1\}$, where $p_\mathcal{R}$ is robot's current position.
\item An edge set $\bar{\mathcal{E}}=\left\{\bar{e}_\alpha: \alpha=0,\ldots |\bar{\mathcal{E}}|-1\right\}$, where each edge $\bar{e}_{\alpha}\equiv\left(\bar{n}_i,\bar{n}_j\right): \bar{n}_i \neq \bar{n}_j, \forall \bar{n}_i,\bar{n}_j \in \bar{\mathcal{N}}$ 
is assigned a cost $w_{ij}$ equal to the shortest collision-free path length computed by applying A$^*$ algorithm~\cite{hart1968formal} on tiling $\mathcal{T}$.
\end{itemize}

Then, the start node of TSP tour is set as the robot's current position $\bar{n}_0$ while the end node $\bar{n}_\beta$ is determined as follows:
\begin{itemize}
    \item Case 1: if $\mathcal{N}_{\widetilde{O}} \neq \emptyset$, then its element that is furthest from $\bar{n}_0$ is set as the end node. This allows for transition to the default greedy coverage strategy upon complete coverage of all \textit{explored} nodes. 
    \item Case 2: if $\mathcal{N}_{\widetilde{O}} = \emptyset$, then the end node is not specified.  
\end{itemize}

Both cases aim to find an open-loop TSP tour starting from the robot's current position and visiting all open nodes. To do this, a dummy node $\bar{n}_{\bar{\mathcal{N}}}$ is added to $\bar{\mathcal{G}}$ to convert this problem into a standard closed-loop TSP that starts and ends at the dummy node. The edge cost between  $\bar{n}_{\bar{\mathcal{N}}}$ and $\bar{n}_0$ is set as $w_{\bar{\mathcal{N}},0}=0$. For case 1, the edge cost between $\bar{n}_{\bar{\mathcal{N}}}$ and the end node $\bar{n}_\beta$ is set as $w_{\bar{\mathcal{N}},\beta}=0$, while for case 2, it is set as $w_{\bar{\mathcal{N}},\beta}=\infty$. The edge costs between $\bar{n}_{\bar{\mathcal{N}}}$ and all the remaining nodes is set as $w_{\bar{\mathcal{N}},i}=\infty, \forall \bar{n}_i\in \bar{\mathcal{N}}\setminus \{\bar{n}_0, \bar{n}_\beta\}$. 

The cost matrix is obtained as       $\mathcal{W}=\left[w_{ij}\right]_{(\bar{\mathcal{N}}+1) \times (\bar{\mathcal{N}}+1)}$. Let $\mathcal{C}$ be the set of all Hamiltonian cycles corresponding to $\mathcal{W}$. Each Hamiltonian cycle $c\in\mathcal{C}$ provides the order in which the nodes are visited and is expressed as follows:
\begin{equation*}\label{eq:ogm_recursive}
c=\big(n(k) \in \bar{\mathcal{N}}, k=0,\ldots,\bar{\mathcal{N}}+1:  n(\bar{\mathcal{N}}+1)=n(0)= \bar{n}_{\bar{\mathcal{N}}}\big)
\end{equation*} 
where $n(k)$ is the node visited at step $k$, and $n(k)\neq n(k')$, $k\neq k'; \forall  k,k'=0,\ldots, \bar{\mathcal{N}}$. The cost of a cycle $c$ is given as:
\begin{equation}
\mathcal{J}\left(c\right)= \sum_{k=0}^{\bar{\mathcal{N}}}w_{n(k)n(k+1)}
\vspace{-0.5em}
\end{equation}
Then, the optimal Hamiltonian cycle $c^{*}$ is
\begin{equation}
c^{*} = \mathop{\argmin}_{c \in \mathcal{C}}\mathcal{J}\left(c\right)
\vspace{-0.5em}
\end{equation}  
The nearest neighbor algorithm~\cite{aarts2003} is used to obtain an initial cycle. Then, the 2-opt algorithm~\cite{aarts2003} is applied over this initial cycle to obtain the optimal solution $c^{*}$. After that, the global coverage tour $\gamma_{global}$ is obtained by removing the dummy node $\bar{n}_{\bar{\mathcal{N}}}$ and robot's current position $\bar{n}_0$ from $c^{*}$. Fig.~\ref{fig:example_part3} shows an example of global coverage tour which covers two \textit{explored} nodes and ends at the \textit{exploring} node. The first element in $\gamma_{global}$ is set as target node $n_{target}$. 

\subsection{Local Coverage Path Planning} \label{localCover}

\subsubsection{Coverage of Exploring Node} 
As shown in Fig.~\ref{fig:example_part1}, the robot covers \textit{exploring} node by following greedy strategy which selects an uncovered cell as target $\tau_{target}$ in the local vicinity based on the priority of movement direction $\{$Left, Up, Down, Right$\}$. These directions are defined with respect to the fixed global coordinate frame. The target cell $\tau_{target}$ is marked as covered. This strategy generates the back-and-forth motions to completely cover target node $n_{target}$. 

\subsubsection{Coverage of Explored Node} 

As shown in Fig.~\ref{fig:example_part4}, the robot covers \textit{explored} node by following the TSP-based optimal path. The start point of TSP tour is set as robot's current position $p_{\mathcal{R}}$ while the end point is set as target cell $\tau_{target}$. To find an open-loop TSP tour to visit all cells of \textit{explored} node, a dummy point is added. The cost between dummy point and $p_{\mathcal{R}}$ (and $\tau_{target}$) is set as 0 while the costs between dummy point and all cells is set as $\infty$. The nearest neighbor algorithm~\cite{aarts2003} and 2-opt algorithm~\cite{aarts2003} are used to solve TSP for the local coverage path $\gamma_{local}$. Then, the robot follows this path to completely cover target node $n_{target}$.

\subsection{Time Complexity Analysis}
The CAP algorithm is summarized in Algorithm~\ref{alg:CAP}. Since environmental map and robot's position are updated continuously during navigation (\textbf{Line 3}), the complexity is evaluated for the steps of subarea identification (\textbf{Line 5}), graph construction (\textbf{Line 7}), global coverage tour planning (\textbf{Line 8}), and local coverage path planning (\textbf{Line 15}).

CAP recursively labels uncovered cells at most once to find all subareas. In the worst case, it searches the entire uncovered area $\mathcal{T}_f^u$ with $O(|\mathcal{T}_f^u|)$ complexity. Next, the graph is updated as follows. First, the current node and associated edges are pruned as its corresponding area is separated into multiple parts $\{\mathcal{A}_\ell\}_{\ell=1}^{L}$. These identified subareas are added as new nodes to the graph with associated edges. In the worst case that the updated environmental map invalidates all old edges, A$^*$ algorithm is applied on $\mathcal{T}_f$ to find new paths for all edges in $\mathcal{E}$, where $|\mathcal{E}|=(|\mathcal{N}|^2-|\mathcal{N}|)/2$. Since A$^*$ algorithm has a complexity of $O(|\mathcal{T}_f|^2)$ and $|\mathcal{T}_f| \gg |\mathcal{N}|$, the overall complexity of graph construction is $O(|\mathcal{T}_f|^2|)$. After that, a TSP of size $\bar{\mathcal{N}}+1$ is formulated and solved for the global coverage tour by the
nearest neighbor algorithm and 2-opt algorithm with $O(\left(\bar{\mathcal{N}}+1\right)^2)$ complexity~\cite{shen2022ct}. Finally, 
the target node $n_{target}$ is covered by using either greedy strategy with  $O(1)$ complexity or TSP-based optimal path with $O(|n_{target}|^2)$ complexity. Since $|\mathcal{T}_f| \geq |\mathcal{T}_f^u| \geq |n_{target}|$ and $|\mathcal{T}_f| \gg |\bar{\mathcal{N}}|$, the overall complexity of CAP is $O(|\mathcal{T}_f|^2)$.

\section{Results and Discussion}
\label{sec:results}
In this section, we present validation results of CAP comparing to five baseline algorithms.

\subsection{Simulation Validations}
\begin{figure}[b]
    \centering
    \vspace{0.6em}
    \includegraphics[width=0.47\textwidth]{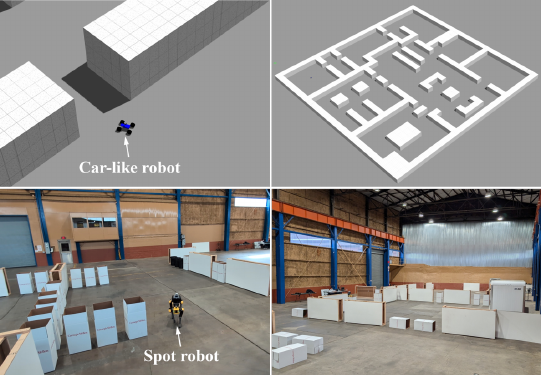}
    \caption{Car-like robot in gazebo scene used for simulation experiments (top row). Spot robot in a large warehouse with configurable obstacles used for real-world experiments (bottom row).}
  \label{fig:warehouse_robot}
\end{figure}

\begin{figure*}[t]
    \centering
    \subfloat[Scene 1: Maze.]{
    \includegraphics[width=0.99\textwidth]{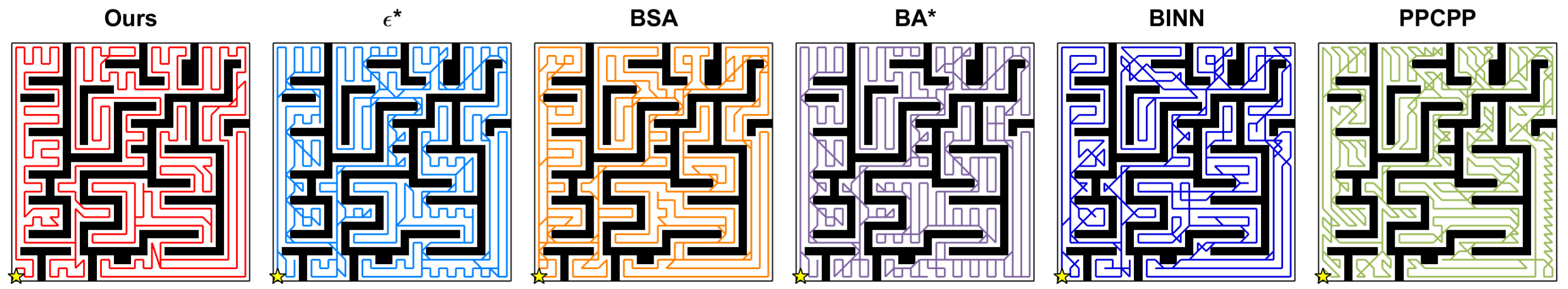}\vspace{-4pt}\label{fig:path_simulation_scenario1}}\vspace{2pt}\\ 
    \centering
    \subfloat[Scene 2: Mall.]{
    \includegraphics[width=0.99\textwidth]{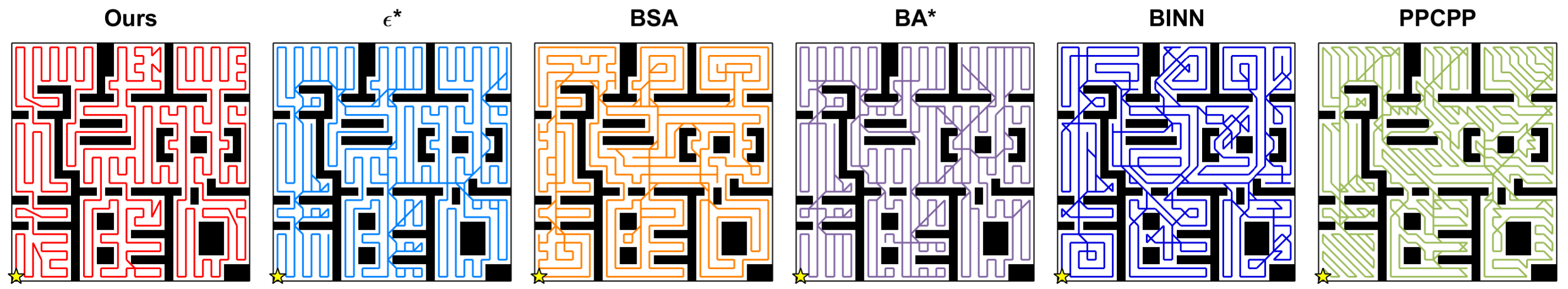}\vspace{-4pt}\label{fig:path_simulation_scenario2}}\vspace{2pt}\\
    \centering
    \subfloat[Scene 3: Office.]{
    \includegraphics[width=0.99\textwidth]{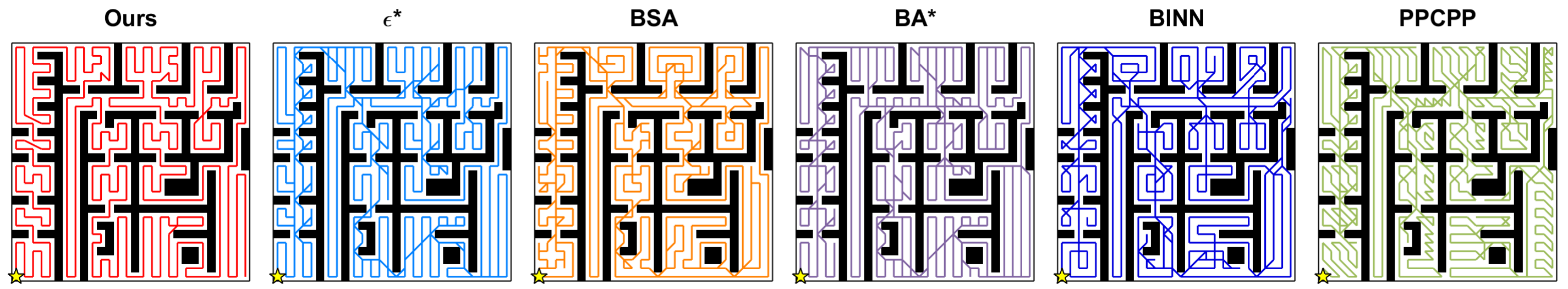}\vspace{-4pt}\label{fig:path_simulation_scenario3}}\\
    \centering
    \subfloat[Scene 4: Warehouse.]{
    \includegraphics[width=0.99\textwidth]{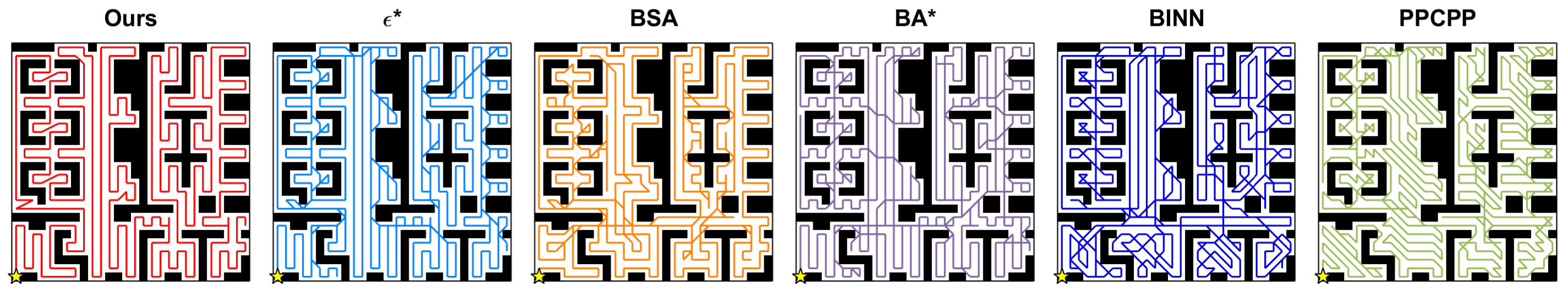}\vspace{-4pt}\label{fig:path_simulation_scenario4}}\\\vspace{5pt}
    \centering
    \subfloat[Comparison of performance metrics.]{
    \includegraphics[width=0.99\textwidth]{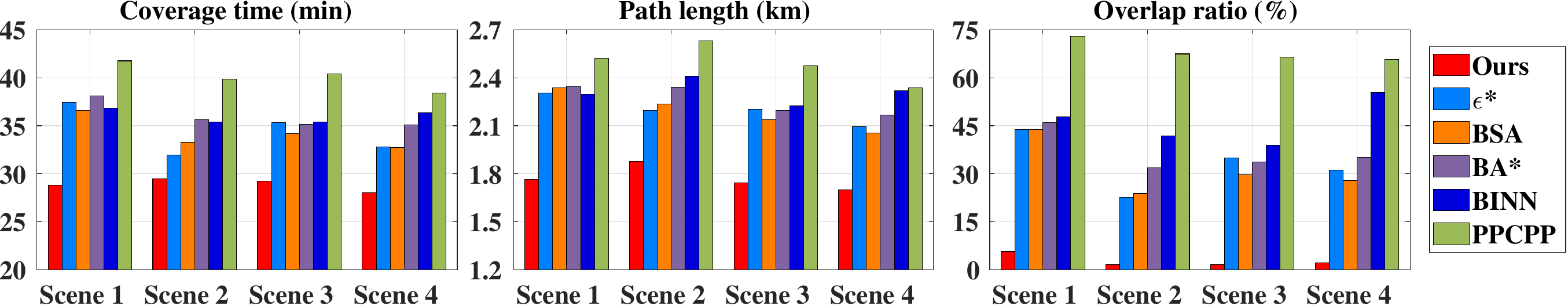}\label{fig:metric_simulation}}\\
    \caption{Performance comparison of CAP with the baseline algorithms in simulations.}
  \label{fig:result_simulation}
  \vspace{-1.0em}
\end{figure*}

The performance of CAP is first evaluated by extensive simulations on Gazebo (Fig. \ref{fig:warehouse_robot}). As shown in Fig.~\ref{fig:result_simulation}, four complex scenes of dimensions $90 m \times 90 m$ are generated with different obstacle layouts and partitioned into a $30 \times 30$ tiling structure for mapping and coverage. Each scene depicts a real situation (e.g., office). A car-like robot is simulated with a maximum speed of $2m/s$ and initialized at the bottom-left corner of the space. It is equipped with a lidar with a detection range of $12m$. Five baseline algorithms including  $\epsilon^*$~\cite{song2018}, BSA~\cite{gonzalez2005bsa}, BA$^*$~\cite{viet2013ba}, BINN~\cite{luo2008bioinspired}, and PPCPP~\cite{hassan2019ppcpp} are selected for performance comparison in terms of coverage time, path length, and overlap ratio $\frac{\bigcup_{t\in[0,T]} \tau(t) \setminus \mathcal{T}_f}{\mathcal{T}_f}$.

Figs.~\ref{fig:path_simulation_scenario1}-\ref{fig:path_simulation_scenario4} show the coverage paths generated by different algorithms in four scenes. As seen, CAP provides complete coverage with less overlapping paths. This is because CAP constructs a coverage guidance graph to capture the global topological structure of the environment, and then computes a global tour on it to improve overall coverage efficiency and also address the issue of bypassing isolated regions. Additionally, it performs TSP-based optimal coverage path within the \textit{explored} subareas to further minimize overlapping paths and local coverage time. In contrast, the baseline algorithms follow greedy strategies to achieve complete coverage without a
global perspective of the entire search area, thus leading to highly overlapping paths and sub-optimal motions. Fig.~\ref{fig:metric_simulation} provides quantitative comparison results in terms of coverage time, path length, and overlap ratio. Overall, CAP achieves significant improvements over the baseline algorithms in all metrics. Particularly, in comparison to the second best baseline algorithm, CAP reduces coverage time by $21\%, 8\%, 15\%,$ and $14\%$ in four scenes, respectively.

\RestyleAlgo{ruled}
\LinesNumbered
\begin{algorithm}[t]
\footnotesize 
$\mathcal{G}$ $\leftarrow$ $n_1$; $\gamma_{global} \leftarrow n_1$; $n_{target} \leftarrow n_1$\;

\While{$\gamma_{global} \neq \emptyset$}{

    $\left\{\mathcal{T}, p_\mathcal{R} \right\}\leftarrow \textbf{UpdateSensorData}()$\;
    
    \If {$\Phi(n_{target})=\widetilde{O}$}
    {
        $\{\mathcal{A}_\ell\}_{\ell=1}^{L}\leftarrow \textbf{IdentifySubarea}(n_{target})$\;

        \If{$L > 1$}
        {
            $\mathcal{G}\leftarrow \textbf{ConstructGraph}(\mathcal{G}, \{\mathcal{A}_\ell\}_{\ell=1}^{L})$\;
            
            $\left\{\gamma_{global}, n_{target} \right\}\leftarrow \textbf{ComputeGlobalTour}(\mathcal{G})$\;
        }
        
    }
    
    \If{$\textbf{isComplete}(n_{target})$}
    {
        $\gamma_{global} \leftarrow \gamma_{global} - n_{target}$\;
        
        $n_{target} \leftarrow \gamma_{global}.\text{top}()$\;
    }
    {
        \textbf{LocalCover}($n_{target}$)\;
    }
    
}
\caption{CAP}
\label{alg:CAP} 
\end{algorithm}
\setlength{\textfloatsep}{0pt}

\begin{figure*}[t]
        \centering
    \subfloat[Ours]{
        \includegraphics[width=0.3\textwidth]{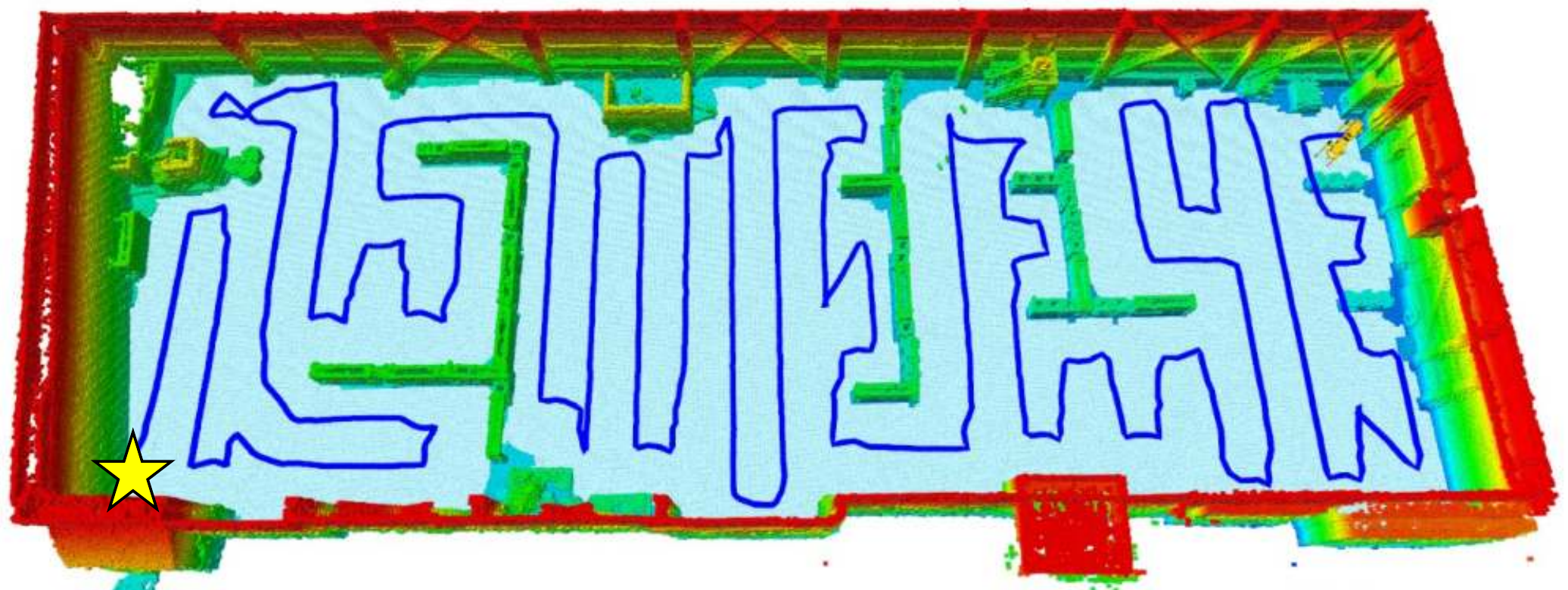}\label{fig:Our_path_experiment_scenario1}\vspace{-7pt}}\quad \hspace{-10pt}
        \centering
    \subfloat[$\epsilon^*$]{
        \includegraphics[width=0.3\textwidth]{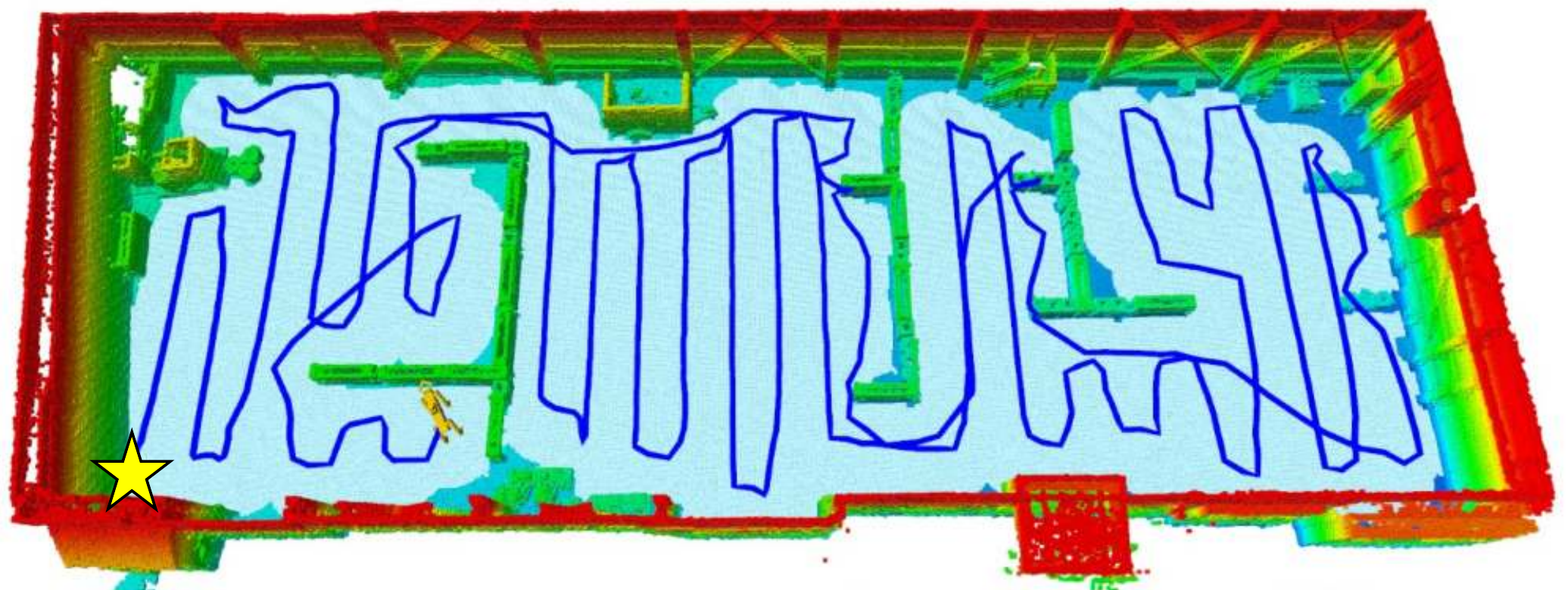}\label{fig:Estar_path_experiment_scenario1}\vspace{-7pt}}\quad \hspace{-10pt}
         \centering
    \subfloat[BSA]{
        \includegraphics[width=0.3\textwidth]{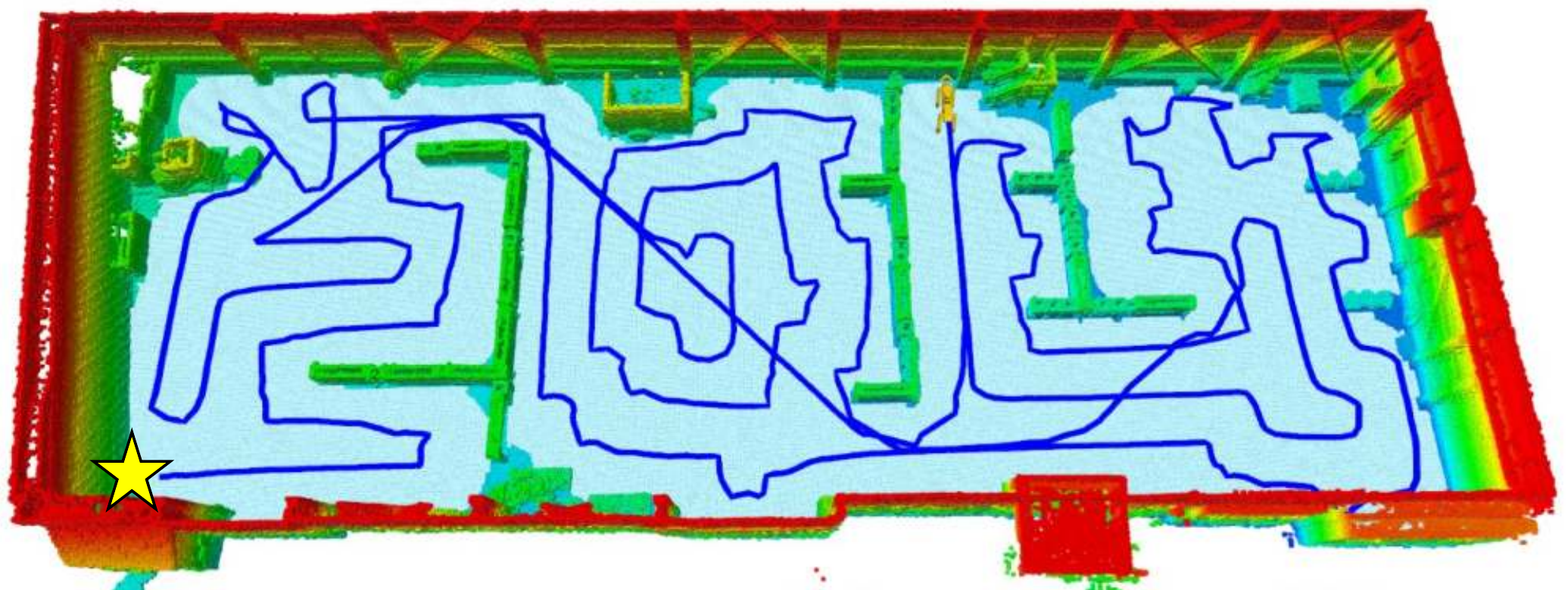}\label{fig:BSA_path_experiment_scenario1}\vspace{-7pt}}\\
        \centering
    \subfloat[BA$^*$]{
        \includegraphics[width=0.3\textwidth]{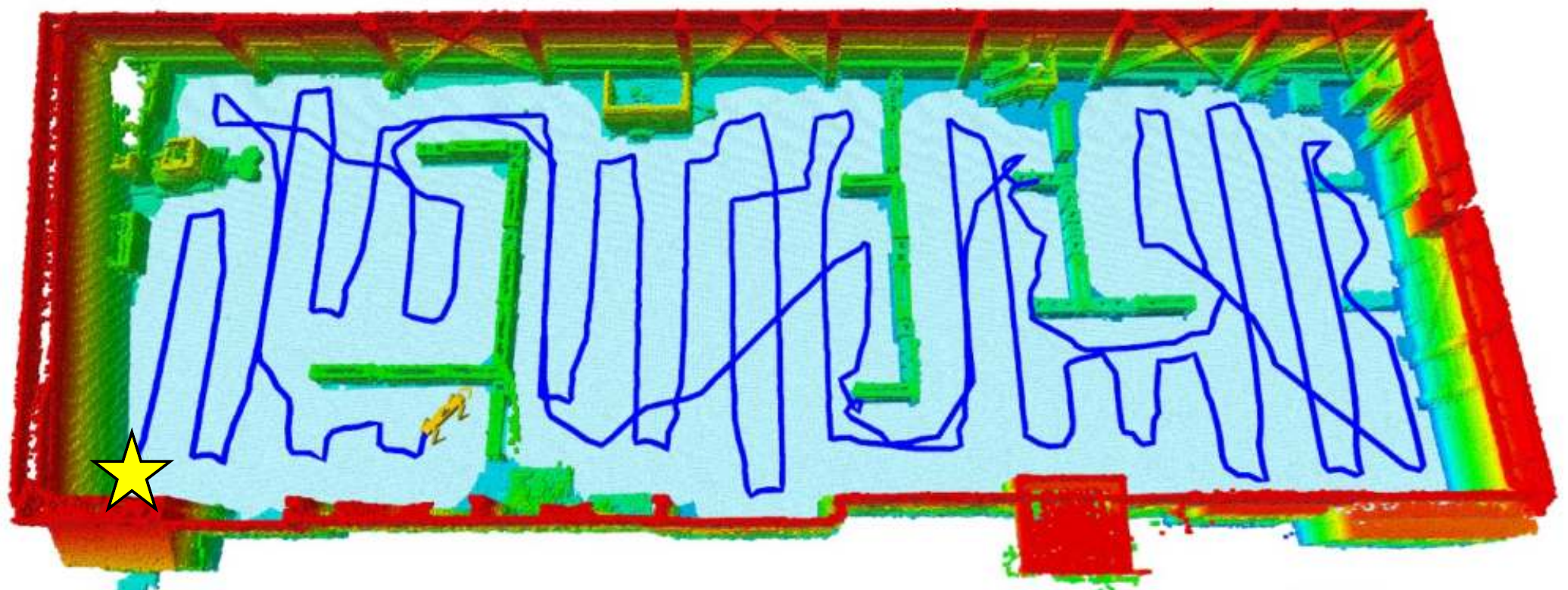}\label{fig:BAstar_path_experiment_scenario1}\vspace{-6pt}}\quad \hspace{-10pt}
         \centering
    \subfloat[BINN]{
        \includegraphics[width=0.3\textwidth]{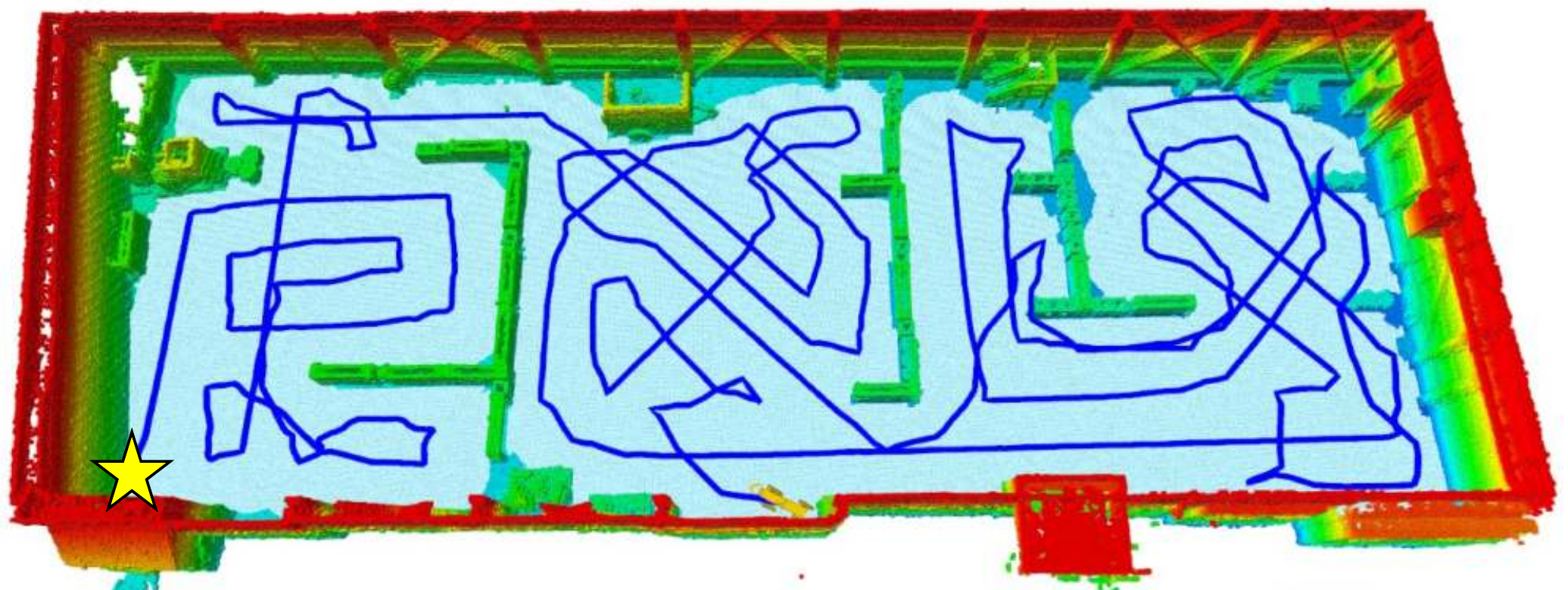}\label{fig:BINN_path_experiment_scenario1}\vspace{-6pt}}\quad \hspace{-10pt}
        \centering
    \subfloat[PPCPP]{
        \includegraphics[width=0.3\textwidth]{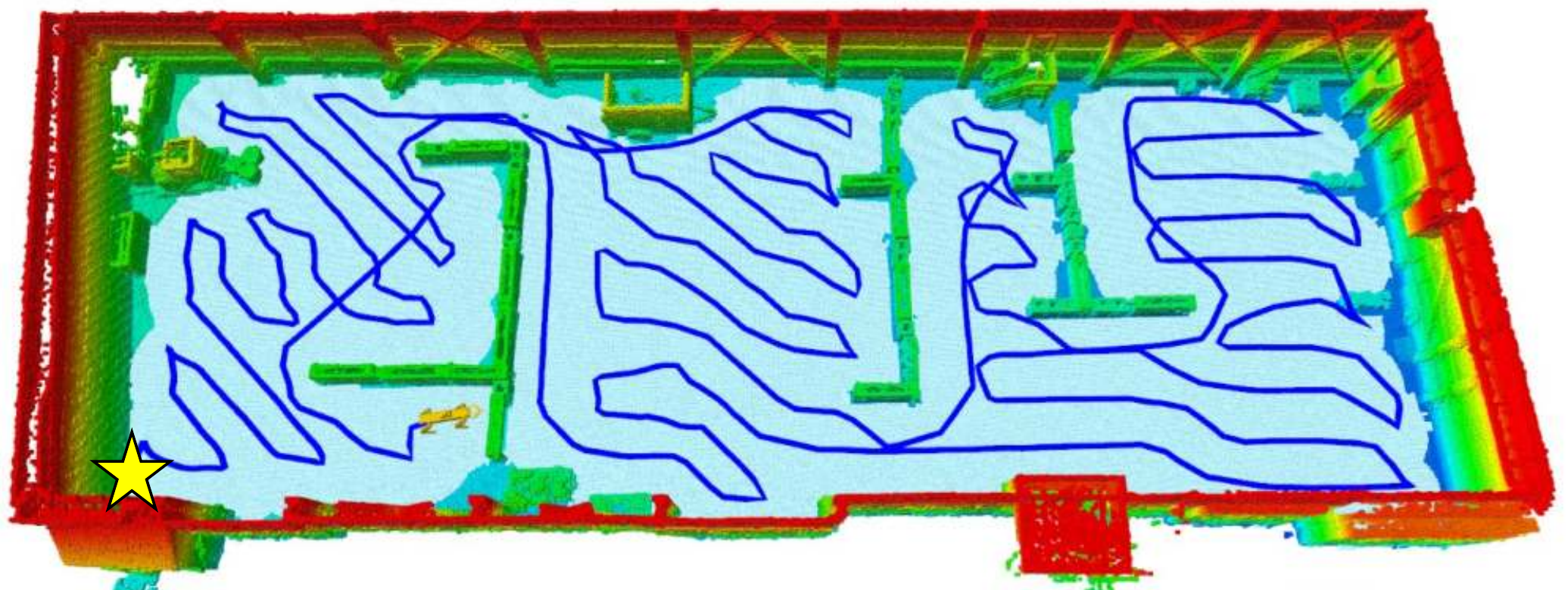}\label{fig:PPCPP_path_experiment_scenario1}\vspace{-6pt}}\\
        \vspace{-4pt}
    \caption{Coverage paths generated by different algorithms in experiment 1.}\label{fig:path_experiment_scenario1} 
    \vspace{-0.5em}
 \end{figure*}

 \begin{figure*}[t]
        \centering
    \subfloat[Ours]{
        \includegraphics[width=0.3\textwidth]{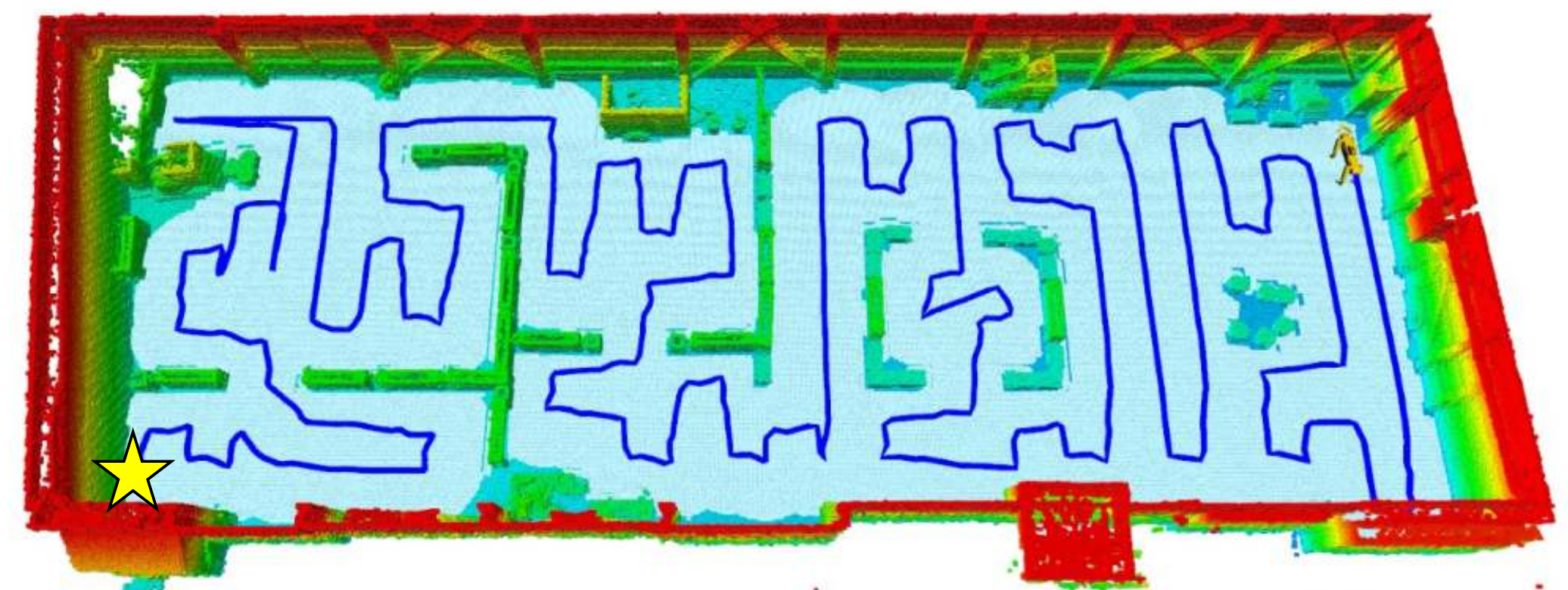}\label{fig:Our_path_experiment_scenario2}\vspace{-6pt}}\quad \hspace{-10pt}
        \centering
    \subfloat[$\epsilon^*$]{
        \includegraphics[width=0.3\textwidth]{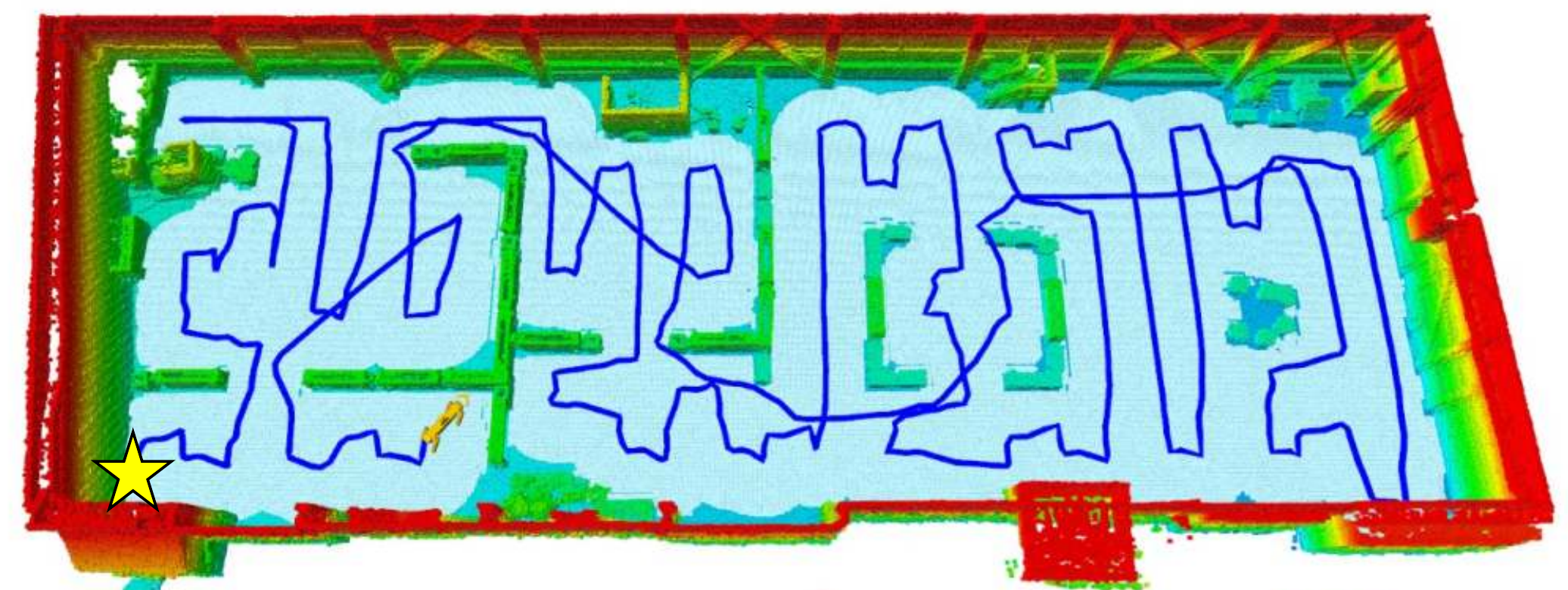}\label{fig:Estar_path_experiment_scenario2}\vspace{-6pt}}\quad \hspace{-10pt}
         \centering
    \subfloat[BSA]{
        \includegraphics[width=0.3\textwidth]{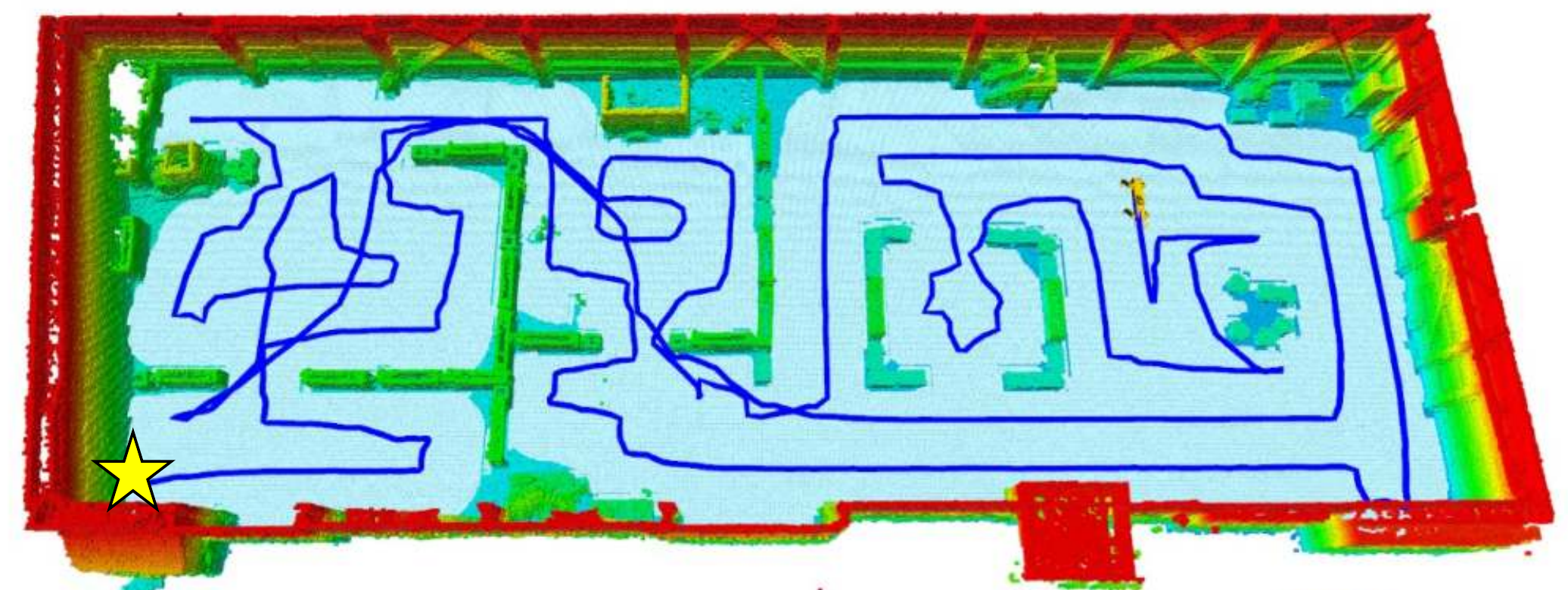}\label{fig:BSA_path_experiment_scenario2}\vspace{-6pt}}\\
        \centering
    \subfloat[BA$^*$]{
        \includegraphics[width=0.3\textwidth]{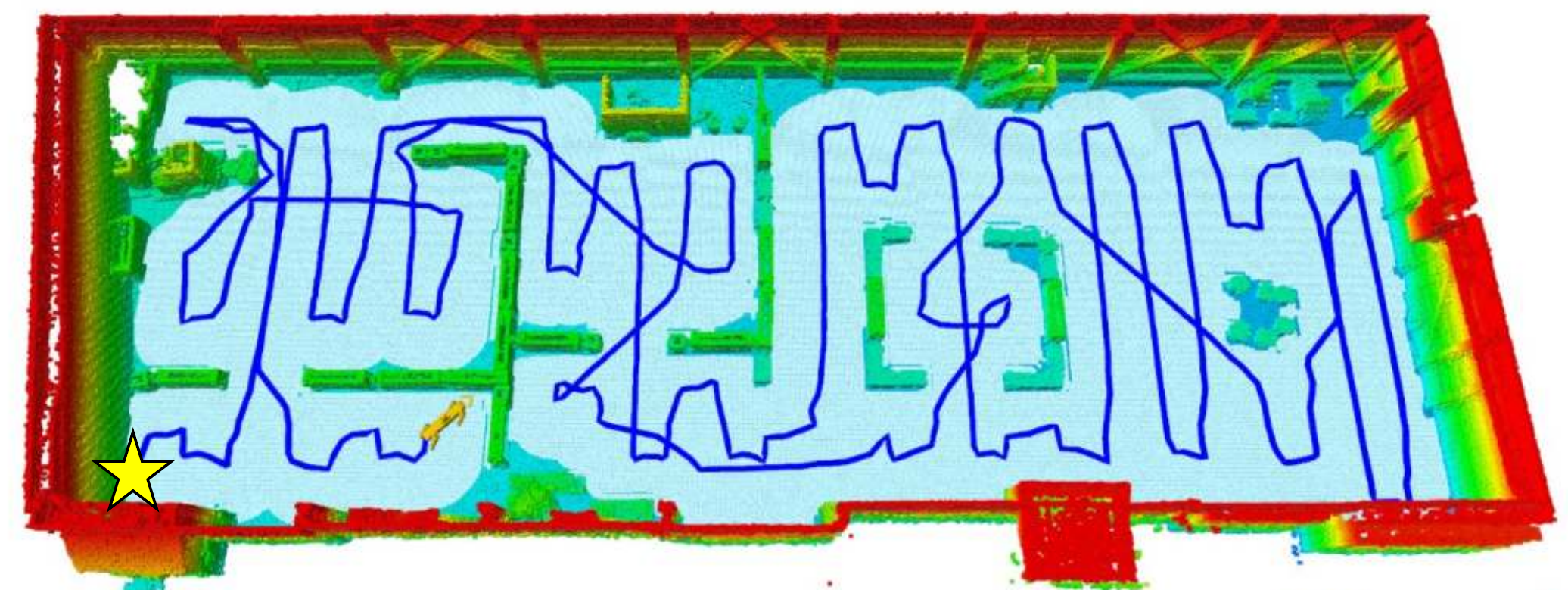}\label{fig:BAstar_path_experiment_scenario2}\vspace{-6pt}}\quad \hspace{-10pt}
         \centering
    \subfloat[BINN]{
        \includegraphics[width=0.3\textwidth]{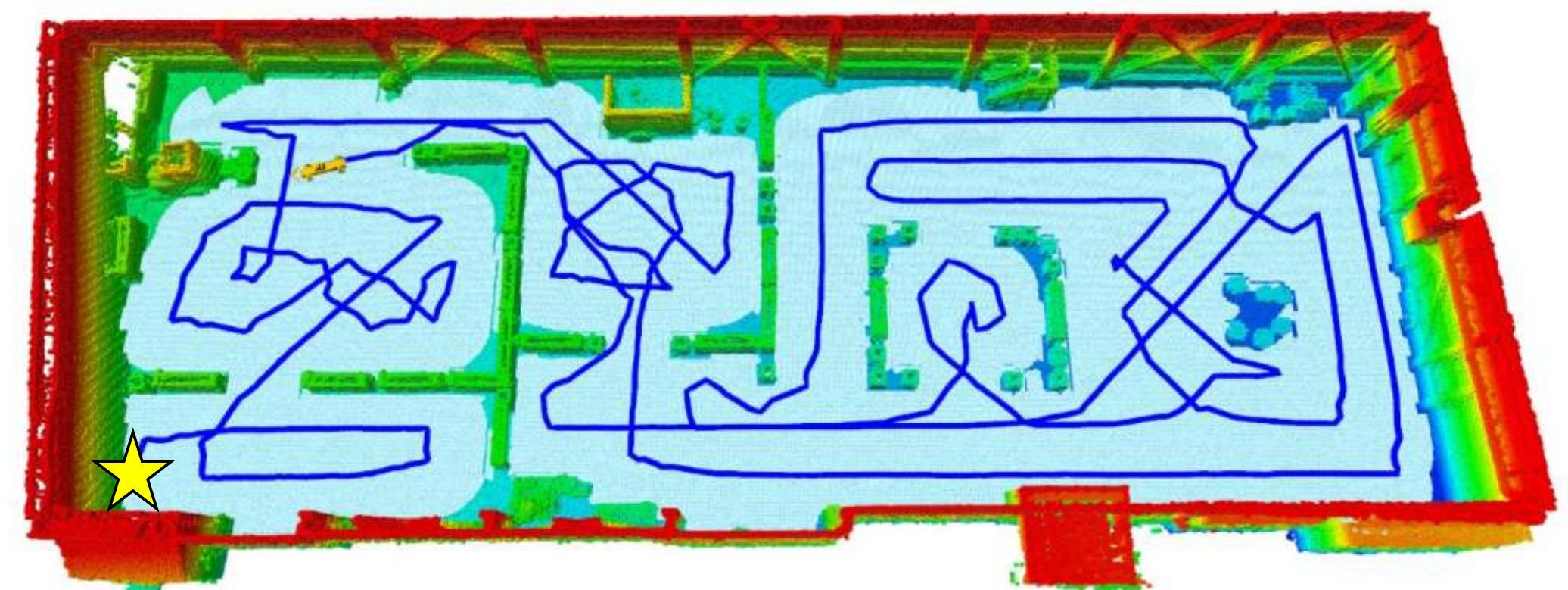}\label{fig:BINN_path_experiment_scenario2}\vspace{-6pt}}\quad \hspace{-10pt}
        \centering
    \subfloat[PPCPP]{
        \includegraphics[width=0.3\textwidth]{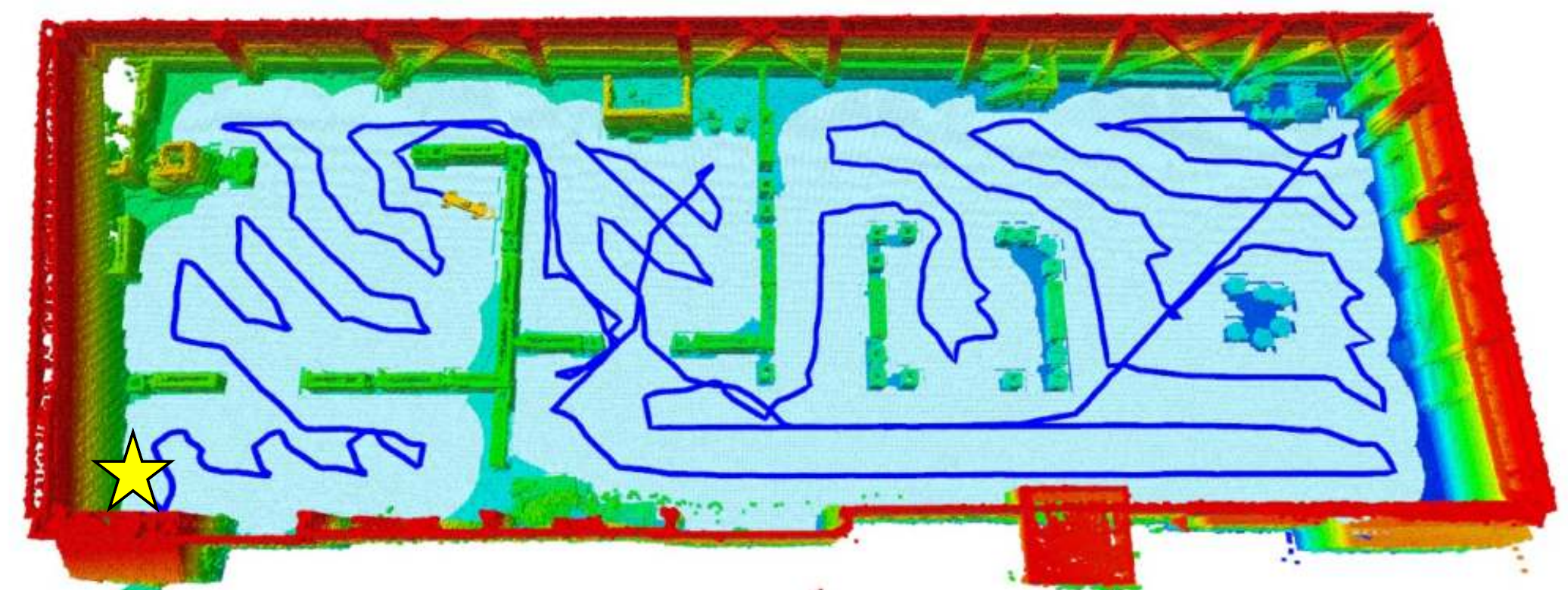}\label{fig:PPCPP_path_experiment_scenario2}\vspace{-6pt}}\\
        \vspace{-4pt}
    \caption{Coverage paths generated by different algorithms in experiment 2.}\label{fig:path_experiment_scenario2} 
    \vspace{-1em}
 \end{figure*}

\vspace{-16pt}

\subsection{Real-World Experiments}
\begin{figure}[t!]
    \centering
    \includegraphics[width=0.48\textwidth]{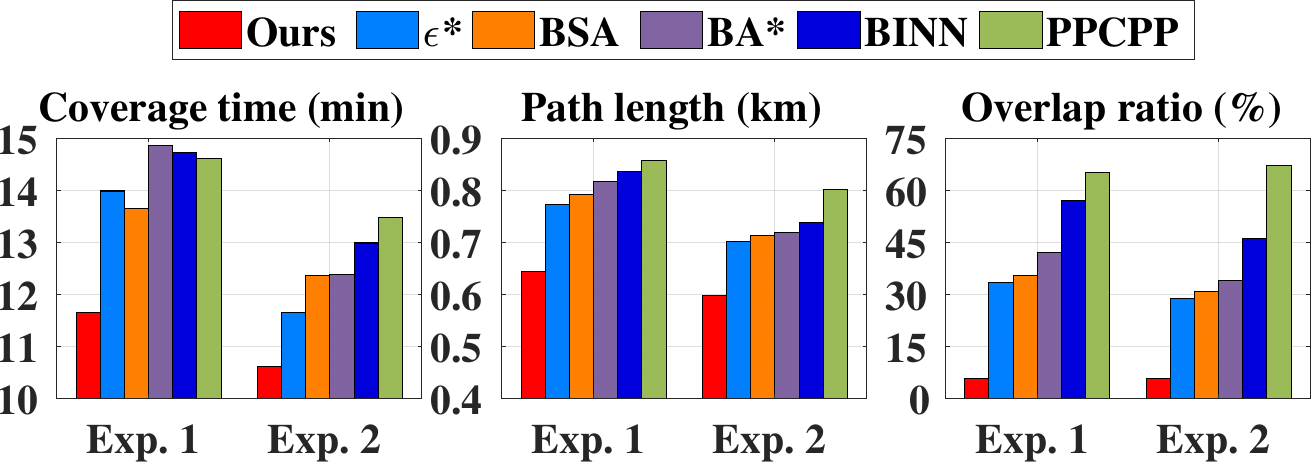}
    \caption{Comparison of performance metrics for two experiments.}
  \label{fig:metric_experiment}
\end{figure}

\vspace{-5pt}
The performance of CAP is also evaluated by two real-world experiments in a $60 m \times 24 m$ warehouse space with different obstacle layouts (Fig. \ref{fig:warehouse_robot}). The space is tiled with $2 m \times 2 m$ cells. We utilize a Boston Dynamics Spot robot equipped with an NVIDIA Jetson AGX Orin and a Velodyne VLP-16 LiDAR. The robot is initialized at the bottom-left
corner of the space. The LiPO~\cite{LiPo} algorithm is used for localization and mapping. An overview of the autonomy stack for this robot is presented in \cite{sriganesh2024systemdesign}. Figs.~\ref{fig:path_experiment_scenario1} and \ref{fig:path_experiment_scenario2} show the coverage paths generated by different algorithms in both the real-world experiments. CAP provides complete
coverage with less overlapping paths. As shown in Fig.~\ref{fig:metric_experiment}, CAP provides a superior solution quality in all metrics. Particularly, in comparison to the second best baseline algorithm, CAP reduces coverage time by $15\%$ and $9\%$ in two experiments.

\section{Conclusions and Future Work} \label{sec:conclusions}

In this paper, we present CAP algorithm for efficient coverage path planning of unknown environments. During online operation, a coverage guidance graph is incrementally built to capture essential environmental information. Then, a hierarchical coverage path is computed on the graph to improve global coverage efficiency and minimize local coverage time. CAP is comparatively evaluated by high-fidelity simulations and real-world experiments. The results show that CAP yields significant improvements in coverage time, path length, and path overlap ratio. Future works include multi-robot coverage with communication limits, coverage of 3D terrain, and coverage of dynamic environments~\cite{Shen_SMART2023}.

\balance
\bibliographystyle{IEEEtran}
\bibliography{reference}

\end{document}